\documentclass[journal]{IEEEtran}
\usepackage[utf8]{inputenc}

\usepackage{graphicx}
\graphicspath{ {figures/} }

\newif\ifusingvcu
\usingvcufalse

\usepackage{cite}
\usepackage{amsmath,amssymb,amsfonts}
\usepackage{mathtools}
\usepackage{algorithm}
\usepackage{booktabs}
\usepackage{algorithmic}

\usepackage{graphicx}
\usepackage{adjustbox}
\usepackage{caption}
\usepackage[caption=false,font=footnotesize]{subfig}
\graphicspath{{Figures}}

\usepackage{tabularx}
\usepackage{pgffor}
\usepackage{hyperref}
\usepackage{soul}

\usepackage[absolute,overlay]{textpos}
\usepackage{xcolor}

\newcommand{\group}[1]{ \left( #1 \right) }

\newcommand{\func}[2]{ { #1 \group{#2} } }

\begin{document}
\title{Self-Supervised and Interpretable Anomaly Detection using Network Transformers}

\author{\IEEEauthorblockN{ Daniel L. Marino \textsuperscript{1},
                           Chathurika S. Wickramasinghe \textsuperscript{1},
                           Craig Rieger \textsuperscript{2},
                           Milos Manic \textsuperscript{1}}\\
	\IEEEauthorblockA{
	\textsuperscript{1} \textit{Virginia Commonwealth University},
	Richmond, VA, USA \\
  \textsuperscript{2} \textit{Idaho National Laboratory},
	Idaho Falls, Idaho, USA \\
	marinodl@vcu.edu, brahmanacsw@vcu.edu, craig.rieger@inl.gov, misko@ieee.org}
	\thanks{
	{
	This work was supported in part by the Department of Energy through the U.S. DOE Idaho Operations Office under Contract DE-AC07-05ID14517, in part by the Resilient Control and Instrumentation Systems (ReCIS) Program of Idaho National Laboratory \\
    This work was supported in part by the Commonwealth Cyber Initiative, an investment in the advancement of cyber R\&D, innovation and workforce development. For more information about CCI, visit cyberinitiative.org.}}
	}

\maketitle

\begin{abstract}

Monitoring traffic in computer networks is one of the core approaches for defending critical infrastructure against cyber attacks.
Machine Learning (ML) and Deep Neural Networks (DNNs) have been proposed in the past as a tool to identify anomalies in computer networks.
Although detecting these anomalies provides an indication of an attack, just detecting an anomaly is not enough information for a user to understand the anomaly.
The black-box nature of off-the-shelf ML models prevents extracting important information that is fundamental to isolate the source of the fault/attack and take corrective measures.
In this paper, we introduce the Network Transformer (NeT), a DNN model for anomaly detection that incorporates the graph structure of the communication network in order to improve interpretability.
The presented approach has the following advantages:
1) enhanced interpretability by incorporating the graph structure of computer networks;
2) provides a hierarchical set of features that enables analysis at different levels of granularity;
3) self-supervised training that does not require labeled data.
The presented approach was tested by evaluating the successful detection of anomalies in an Industrial Control System (ICS).
The presented approach successfully identified anomalies, the devices affected, and the specific connections causing the anomalies, providing a data-driven hierarchical approach to analyze the behavior of a cyber network.

\end{abstract}


\section{Introduction}

\begin{textblock*}{5cm}(1.4cm,26.7cm) 
   Preprint. Under review.
\end{textblock*}

The use of Information and Communication Technologies (ICTs) in Industrial Control Systems (ICSs) has been driven by the need for better efficiency and connectivity \cite{marino2021access}.
However, ICTs have also opened the door to cyber criminals, harming security and resilience \cite{en14051409}.
Network traffic monitoring is one of the core technologies for maintaining a secure and reliable network.
As such, it is a valuable asset to secure critical infrastructure that relies on ICTs.

\begin{figure}[t]
\centering
\adjustimage{max size={0.99\linewidth}{0.7\paperheight}}{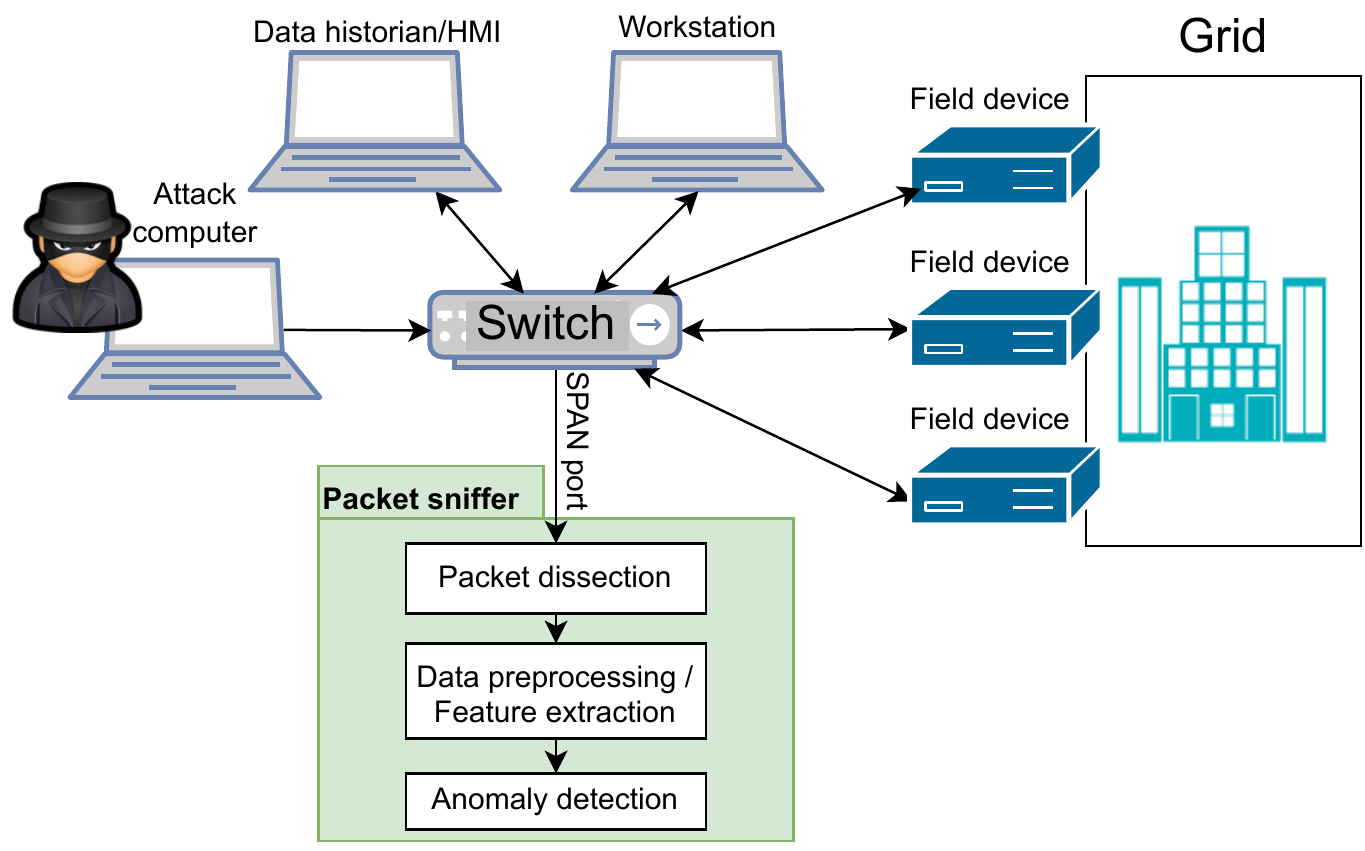}
\caption{Anomaly detection in network traffic monitoring}
\label{fig:CyberADS}
\end{figure}

Figure \ref{fig:CyberADS} shows an example of using ML for network traffic monitoring in an ICS network \cite{marino2019cyber,marino2021access,marino2021data}.
In this example, a packet-sniffer is connected to a switch in order to monitor the communication between devices in the network.
The sniffer is connected to a switch port analyzer (SPAN port).
All incoming and outgoing communication passing through the switch is mirrored to the SPAN port, allowing the packet sniffer to have access to all packets communicated through the switch. 
The data acquired by the sniffer is analysed in order to find anomalies. 
The data is processed first by performing packet dissection.
A rolling window is used to analyze sections of the data, extracting a series of manually engineered features to be used as inputs to ML anomaly detection systems.
The normal behavior of the system is learned by ML algorithms such that any behaviors that are different from previously seen data are flagged as anomalies.

Existing approaches demonstrate the capability of ML models to identify abnormal cyber behavior \cite{marino2019cyber,marino2019data,zhao2021lstm,skopik2020synergy}.
However, these approaches often lack the interpretability of the results as they only report an anomaly without any further information about the source or cause of the anomaly.
This is an issue, specially if the approach is applied in monitoring applications.
Reporting the detection of an anomaly alone does not provide enough information for an operator to isolate the source of the problem and plan corrective measures.
Many off-the-shelf ML models used for anomaly detection follow a black-box design, where the inner workings of the model are usually not understood by the user \cite{gunning2017explainable}.
Running these models without taking advantage of the structure in the data leads to approaches that obfuscate exploratory analysis.
Designing an approach that not only detects anomalies but also aids on exploratory analysis is fundamental for improving interpretability.
Preserving the graph structure of the problem provides a representation that opens the door for better interpretability and diagnosis.

This paper presents the Network Transformer (NeT), a ML model that uses graph representations to improve the interpretability of anomaly detection in network traffic monitoring.
The presented model allows to identify anomalies while also providing a series of hierarchical features that allow to identify devices affected by the anomalies and the connections responsible for the anomalies.
The approach leverages self-supervised learning to train the model using unlabeled data.
A multi-processing pipeline is presented as a prototype for scalability to large datasets.
The rest of the paper is organized as follows:
section \ref{section:network_transformer} describes the presented Network Transformer approach;
section \ref{section:cyber_experiments} presents the experimental evaluation;
section \ref{section:cyber_related_work} presents related work;
section \ref{section:cyber_conclusion} concludes the paper.

\section{Interpretable Anomaly Detection: Network Transformer}
\label{section:network_transformer}
\begin{figure}
    \centering
    \includegraphics[scale=0.7]{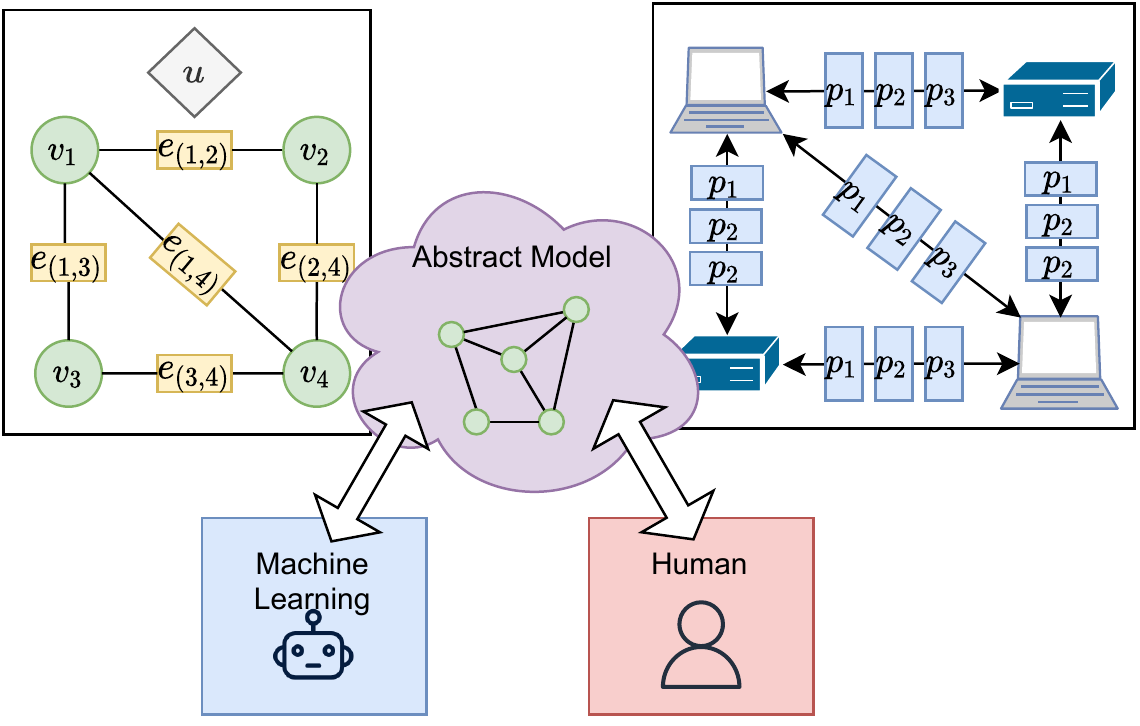}
    \caption{A graph model allows to provide an abstract representation that is shared between human and machine}
    \label{fig:abstractgraphmodel}
\end{figure}

In order to create an interpretable model for anomaly detection, we design an approach that leverages the graph structure of computer networks.
As shown in Figure \ref{fig:abstractgraphmodel}, the idea is to have an abstract model of the problem that is shared by the human and the machine learning model.
The abstract model in this case is the graph representation of the computer network being monitored.
By embedding this abstract model into the ML approach, we are able to leverage our understanding of the system as a graph in order to extract useful information about the anomalies detected in the system.

\begin{figure*}[t]
    \centering
    \includegraphics[scale=0.7]{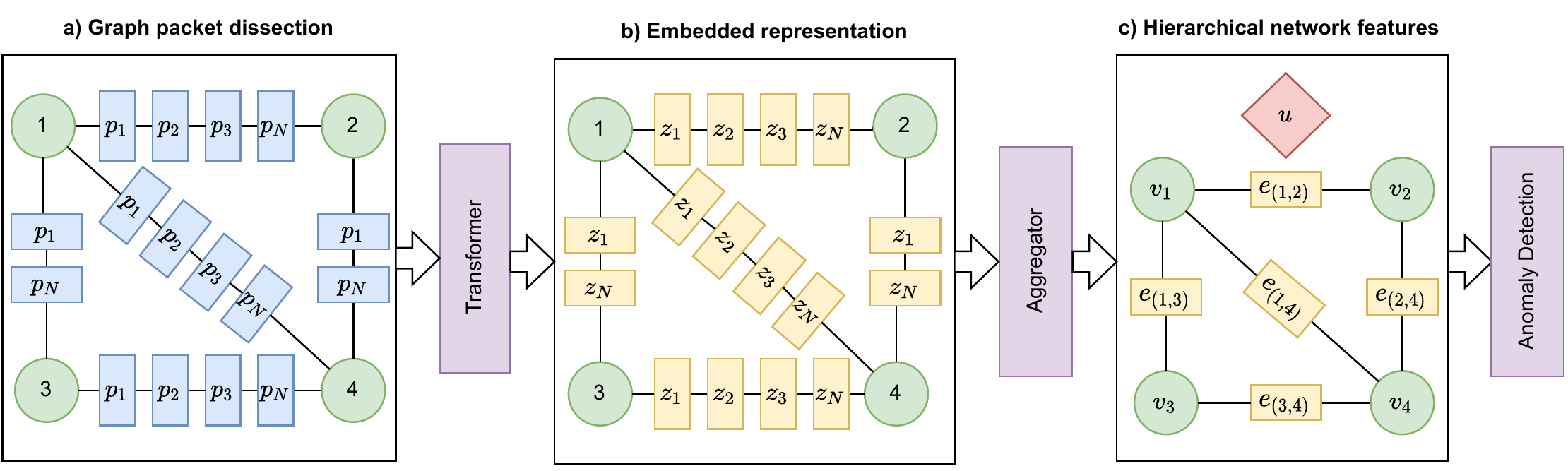}
    \caption{{
    Diagram of the presented Network Transformer (NeT) model. The model allows the extraction of a series of hierarchical network features to represent the monitored computer network as a graph.
}}
    \label{fig:NetworkTransformerOverview}
\end{figure*}

Figure \ref{fig:NetworkTransformerOverview} shows the overview of the presented Network Transformer (NeT) approach.
The approach consists of: a) a packet dissector that groups packages by their respective source and destination addresses; b) an embedded representation obtained using a Transformer Neural Network; c) an Aggregator that extracts a series of hierarchical network features that represent the communication graph.
The features extracted by the NeT model are fed to anomaly detection algorithms to identify anomalies at different levels of granularity.
The approach uses the IP address of the devices in the network to represent the nodes in a graph. 
The packets communicated between the devices are used to represent the edges of the graph.
The hierarchical network features are ultimately used for anomaly detection.
The hierarchical features include global features representing the entire graph, node features representing each node, and edge features representing individual connections.
The approach uses Transformer Neural Networks \cite{NIPS2017_Transformer}, which are currently the state-of-the-art ML model for sequence modeling tasks.
The hierarchical network features introduced in this paper are inspired by Graph Networks \cite{graphnets2018arxiv}.
The following subsections will expand the description of each one of the aforementioned components.

\subsection{Graph packet dissection}
\label{section:graph_dissection}
The approach starts by dissecting packet windows in order to identify the source and destination addresses. 
Packets are grouped by their respective source-destination pairs.
A graph is constructed where each node represents an IP address.
Edges consist of the list of packets communicated between nodes.
Each packet is dissected in order to create an initial feature representation that is suitable for a ML model.
We considered two sets of features: TCP features and raw Byte features.
TCP features are presented in table \ref{tab:TCPFeatures}.
Raw Byte features are presented in table \ref{tab:ByteFeatures}.

\begin{table}[t]
    \centering
    \scalebox{\ifusingvcu 0.75 \else 0.95 \fi}{
    \begin{tabular}{c|c|p{0.63\linewidth}}
         \textbf{Type}        & \textbf{Feature} & \textbf{Description} \\
         \toprule
         Binary      & direction   & Direction of the packet: 0 for $i$ to $j$, 1 for $j$ to $i$.  \\
         Binary      & tcp syn     & TCP SYN flag             \\
         Binary      & tcp ack     & TCP acknowledgment flag. \\
         Binary      & tcp psh     & TCP push function.       \\
         Binary      & tcp urg     & TCP urgent flag.         \\
         Categorical & protocol       & Layer of communication: ARP, IP, IPv6, TCP, or UDP.  \\
         Categorical & service     & Service (inferred from port used): DNP3, ftp, http, git, telnet, ssh, x11, etc.  \\
         Numerical   & len         & Size (number of bytes) of the packet.         \\
         Numerical   & delta time  & Difference in seconds between current packet and previous packet.     \\
         Numerical   & source port & Port used by the source device.      \\
         Numerical   & dest. port  & Port used by the destination device.  \\
         Numerical   & tcp seq     & TCP sequence number.         \\
         Numerical   & tcp ttl     & TCP time to live.         \\
         Numerical   & tcp window  & TCP window.         \\
         Numerical   & port 22 len & Size of the packet (bytes) when using port 22. \\
         Numerical   & port 23 len & Size of the packet (bytes) when using port 23.  \\
    \end{tabular}
    }
    \caption{Packet features when using TCP dissection.}
    \label{tab:TCPFeatures}
\end{table}

\begin{table}[t]
    \centering
    \scalebox{0.9}{
    \begin{tabular}{c|c|p{0.63\linewidth}}
         \textbf{Type}        & \textbf{Feature} & \textbf{Description} \\
         \toprule
         Binary      & direction   & Direction of the packet: 0 for $i$ to $j$, 1 for $j$ to $i$.  \\
         Categorical & protocol    & Layer of communication: ARP, IP, IPv6, TCP, or UDP.  \\
         Numerical   & len         & Size (number of bytes) of the packet.         \\
         Numerical   & delta time  & Difference in seconds between current packet and previous packet.     \\
         Numerical   & bytes       & List of bytes extracted from the packet (0-512).  \\
    \end{tabular}
    }
    \caption{Packet features when using raw Bytes.}
    \label{tab:ByteFeatures}
\end{table}

\subsection{Embedded packet representations, Transformer, and self-supervised training}
\begin{figure}[t]
    \centering
    \includegraphics[scale=0.68]{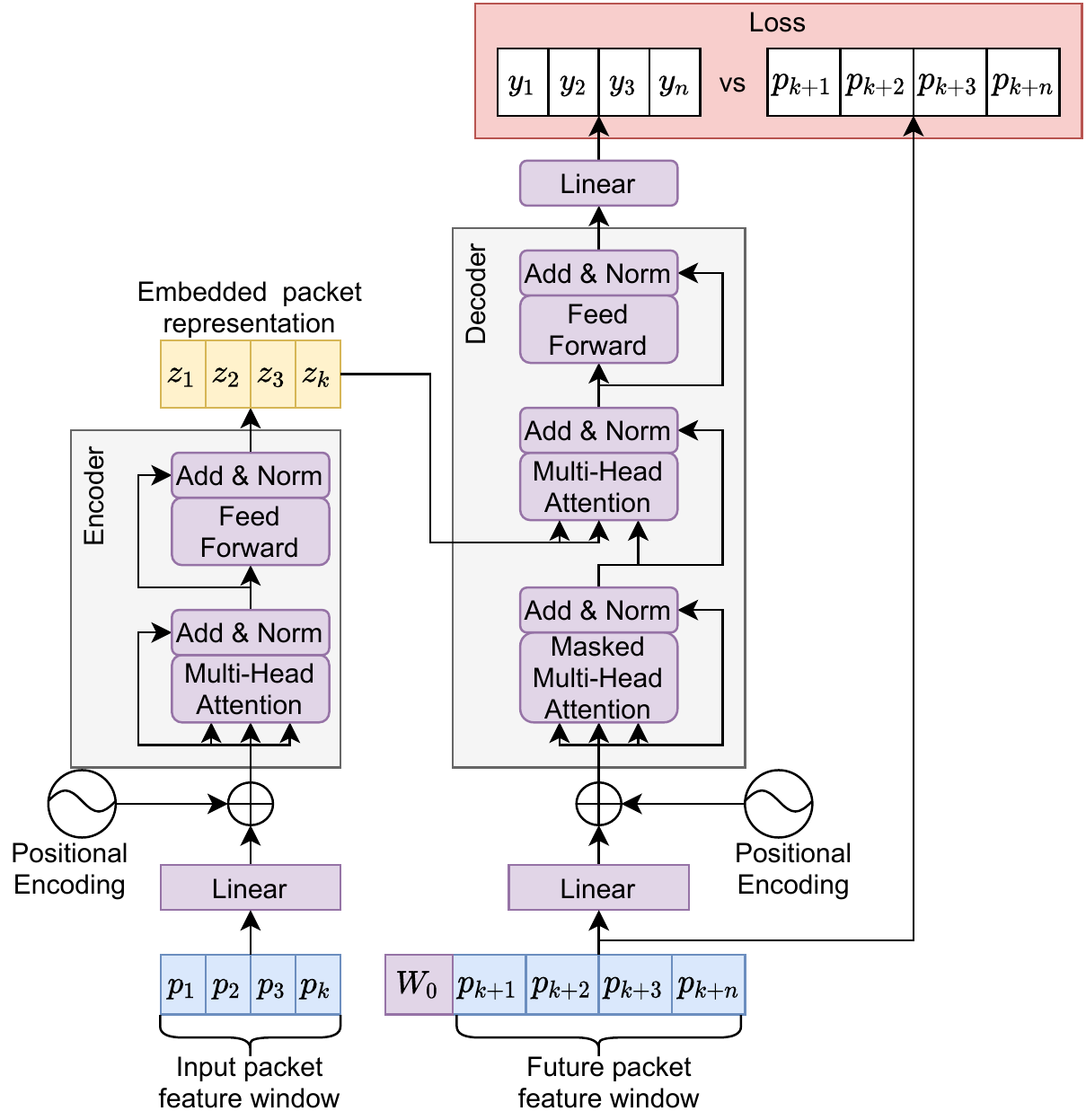}
    \caption{Transformer model and self-supervised training.}
    \label{fig:transformer}
\end{figure}

The Transformer Neural Network model is presented in Figure \ref{fig:transformer}.
We use the Transformer as it is the current state-of-the-art model for sequence modeling.
The Transformer uses attention layers that allow capturing long-range dependencies directly \cite{wang2018non}.
Attention layers also improve the efficiency of the training algorithm as sequences can be processed in parallel; this is in contrast to LSTM and GRU models whose inherently sequential nature precludes parallelization \cite{NIPS2017_Transformer}.
We use a modified version of the Transformer model presented in \cite{NIPS2017_Transformer}, which was originally used to train language models.
Like a sentence is composed of a sequence of words, the edges in our network are composed of a sequence of packets. 
Thus, we use the Transformer to encode the list of packets $p_k$ in each edge to a latent representation $z_{(i,j)}$.

The Transformer model used in this work is trained using a self-supervised learning approach \cite{liu2021self}.
The Transformer is trained to predict the next $n$ packets (unobserved) given the sequence of past $k$ packets (observed).
As shown in Figure \ref{fig:transformer}, the Transformer consists of an Encoder-Decoder network.
The Encoder network encodes the input packets into a set of embedded representations that are used by the decoder to predict a series of future packets.
This approach allows us to leverage unlabeled data as the input-output packet sequence can be extracted by dividing unlabeled packet windows into two.
Once the Transformer is trained, we use the Encoder network to extract packet features $z_n$ from the edges of the graph \ref{section:graph_dissection}.

As shown in Figure \ref{fig:transformer}, the model is composed of four types of layers: 

\begin{itemize}
    \item Linear: applies an affine transformation $f(x) = Wx + b$ of the input using a weight matrix $W$ and a bias vector $b$.
    \item Feed Forward: applies an non-linear transformation using stacked layers $f(x) = \sigma(Wx + b)$ where $\sigma$ is an activation function. We use the ReLU activation function.
    \item Add \& Norm: this layer adds a residual connection \cite{he2016deep} followed by Layer Normalization \cite{ba2016layer}. 
    \item Multi-Head Attention: this layer consists of several scaled dot-product attention models that evaluate the input in parallel.
    The scaled dot-product attention is a function that performs an evaluation of a query matrix ($Q$) over a series of key ($K$) value ($V$) matrices:
    \begin{align*}
        \text{Attention}(Q,K,V) = \func{\text{softmax}}{\dfrac{Q K^T}{\sqrt{d_z}}} V
    \end{align*}
    where $d_z$ is the dimension of the encoded feature representations, which is specified when instantiating the Transformer model. 
\end{itemize}

Figure \ref{fig:transformer} shows $W0$, which is a vector used at the beginning of the target sequence to shift the position of the packets on the right.
Shifting the input to the right allows the Decoder to predict the next packet given the series of previously predicted symbols, making the Transformer model auto-regressive \cite{NIPS2017_Transformer}.
In our implementation, the vector $W0$ is a trainable parameter that is learned when training the Transformer model.
The position encoding in Figure \ref{fig:transformer} uses the sine and cosine functions presented in \cite{NIPS2017_Transformer}.

The loss function in Figure \ref{fig:transformer} is used to evaluate the performance of the Transformer network to predict the future window of packets.
As shown in Tables \ref{tab:TCPFeatures} and \ref{tab:ByteFeatures}, the packets are represented by a series of binary, categorical, and numerical values.
We use this distinction to evaluate the loss depending on the type of the predicted value.
Numerical values use a quadratic loss, while binary and categorical values use a cross-entropy loss.

\subsection{Hierarchical Graph Features}
\label{section:graph_features}

\begin{figure}[t]
    \centering
    \includegraphics[scale=0.75]{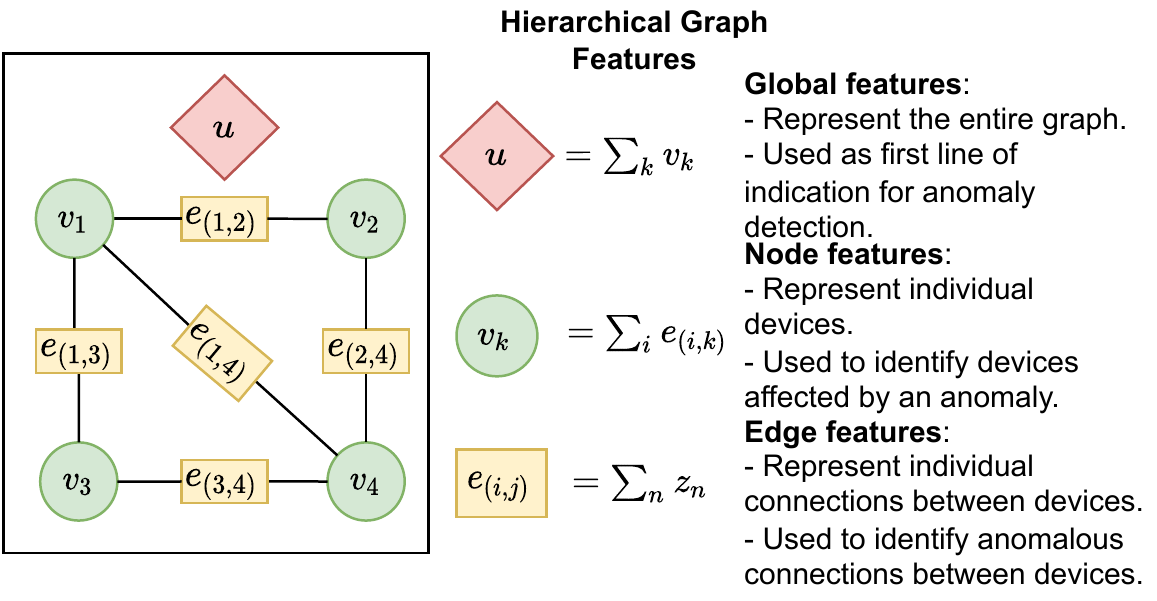}
    \caption{Hierarchical graph features: Global, Node, and Edge features.
    }
    \label{fig:graph_features}
\end{figure}

Figure \ref{fig:graph_features} shows the hierarchical graph features used to represent the monitored network.
The graph features are computed on encoded packet windows obtained with the Transformer encoder.
A packet window is a list of packets obtained during a specified time window (30 seconds in this paper).
The hierarchical graph features consist of: global, node, and edge features.
The specification of the features is inspired by Graph Nets \cite{graphnets2018arxiv}.
To compute the features, we start with encoded packet features $z_n$, which are obtained using the Encoder from the Transformer model (see Fig. \ref{fig:transformer}).
Edge features $e_({i,k})$ represent individual connections between devices.
The value of $e_({i,k})$ is obtained by summing the list of encoded packets $z_n$ communicated between nodes $i$ and $k$.
Node features are obtained by summing the Edge feature values $e_{(i,k)}$ attached to the node.
Global features $u$ are obtained by summing all Edge features in the graph.

The graph features represent the computer network in a given time window.
They represent the network in different layers of abstraction, using a format that is easy to interpret and exploit in order to extract information.
We train anomaly detection algorithms for each level of abstraction (global, nodes, edges) in order to detect anomalies and extract information in different levels of granularity.
The algorithms are trained using only data collected during the normal operation of the system.
In this paper, we consider three anomaly detection algorithms: Local Outlier Factor (LOF), One-Class SVM (OCSVM), and Autoencoders (AE).

Global features provide the highest level of abstraction by representing the entire network for a given window of packets.
Global features provide a way to identify anomalies effectively without dealing directly with individual nodes or connections.
This serves as the first indication of anomalous behavior that considers the network as a whole.

Node features are used to characterize the behavior of individual devices.
They provide the next level of abstraction after global features.
After an anomaly is detected with the global features, node features are used to identify the devices involved in the anomalous event.
Edge features provide the next level of abstraction after Node features.
Edge features allow us to identify exactly which connections are exhibiting anomalous behavior.
{  Even when we are using a complex DNN Transformer model, 
the hierarchical graph features provide a structure that is easy to understand. 
This structure can be exploited in order to extract information such as devices and connections involved in an anomaly.
}

\subsection{Scalability}
\begin{figure*}[t]
    \centering
    \includegraphics[scale=0.75]{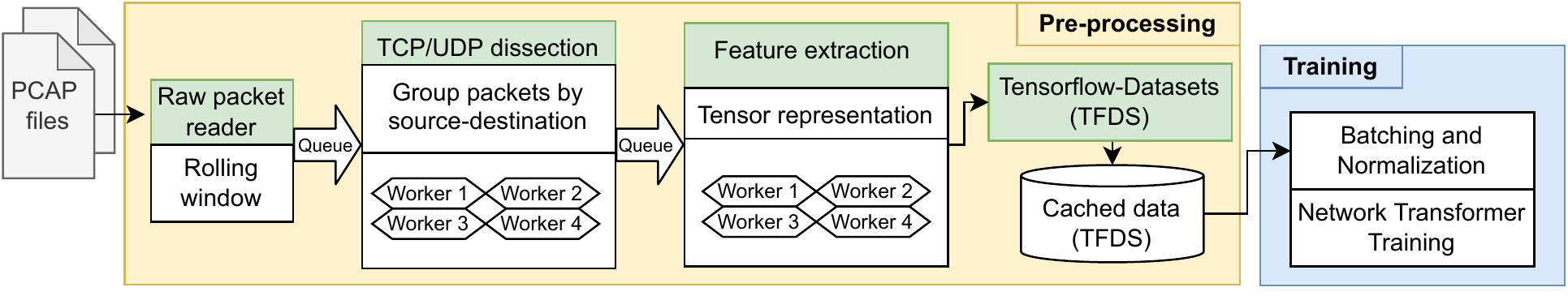}
    \caption{Data pre-processing pipeline}
    \label{fig:PipelineScalability}
\end{figure*}

Network packet data is often characterized by a very high volume of samples.
Attacks such as network scans and DOS often flood the computer network with a high volume of packets. 
As a result, ML algorithms need to be designed in order to handle large quantities of data.
Performing the graph packet dissection and training of the Network Transformer required the development of a multiprocessing pipeline in order to scale to these types of datasets.

Figure \ref{fig:PipelineScalability} shows the developed multiprocessing pipeline used to train the Network Transformer.
The pipeline pre-processes a series of packet capture files (PCAP) into a series of features formatted as tensors (multi-dimensional arrays), ready to be consumed by Tensorflow.
The first stage consists of extracting windows of consecutive raw packets using a rolling window.
In our case, we used a window of 30 seconds, but this can be tuned depending on the application.
The second processing stage consists of TCP/UDP dissection, which groups packets by the source-destination IP address.
The third processing stage extracts features according to tables \ref{tab:TCPFeatures} and \ref{tab:ByteFeatures}.
These features are grouped using the source-destination address and then packaged in a tensor representation.
These tensor feature representations are then managed by Tensorflow-Datasets, a library that creates a cache of the pre-processed values in order to be consumed efficiently when training the Network Transformer.
The training uses Stochastic Gradient Descent, more specifically the ADAM algorithm \cite{kingma2014adam}, which scales to very large datasets.
Once the NeT is trained, the features can be used for downstream ML operations without re-training the model.
The pre-processing pipeline in this work was implemented using Python multiprocessing library, but the same approach can be horizontally scaled with libraries such as Hadoop or Spark.

\section{Experiments}
\label{section:cyber_experiments}
This section presents the experimental analysis of the Network Transformer.
We used packet captures from a real ICS system to evaluate the performance of the presented Network Transformer approach.
We evaluate the performance on an anomaly detection task and the ability of the model to provide an indication of devices and connections related to an anomaly.

\subsection{Data Collection}
\label{section:cyber_data_collection}

\begin{figure}[t]
    \centering
    \includegraphics[scale=0.7]{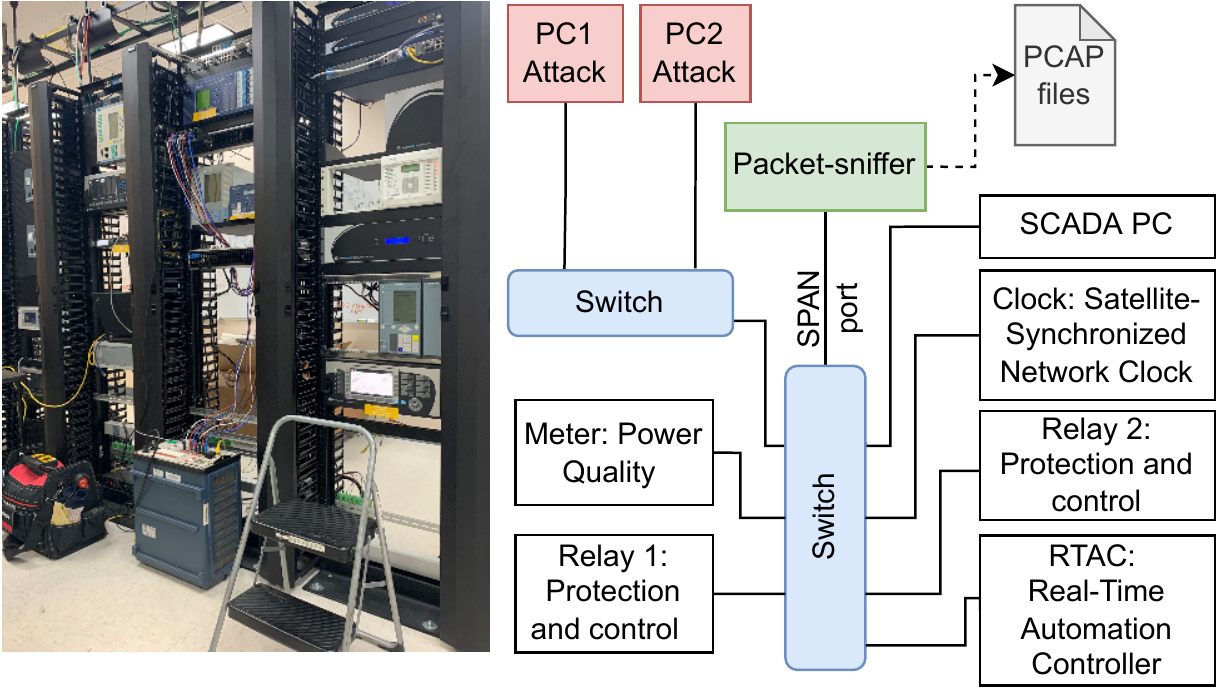}
    \caption{
    ICS testbed used to collect data. Testbed and data collection operated by POWER Engineers Inc.
    }
    \label{fig:SCADAdata}
\end{figure}

The presented approach was evaluated using a dataset of packet captures collected in an ICS.
The data was provided by Idaho National Laboratory (INL).
The ICS network is presented in Figure \ref{fig:SCADAdata}.
The network is composed of two attack computers, two protection relays, one power quality meter, one real-time automation controller (RTAC), one satellite synchronized network clock, and one SCADA PC.
All devices are connected using two network switches.
The RTAC is using the DNP3 protocol to communicate with the relays, the meter, and the SCADA PC.
A sniffer is connected to a SPAN port in the switch. The SPAN port forwards all the packet data passing through the switch to the sniffer, which stores the data in PCAP files for later analysis.
In total, the collected data consisted of 6.036.046 packets.

For experimental evaluation, five types of scenarios were considered:

\begin{itemize}
    \item Normal scenario: Consist of normal operation of the system without any cyber disturbance/attack executed.
    \item Flood attack: Launches a flood of packets using hpin3 command with random source IP address. The attack is launched from the PC1 attack device. The attack targets Relay 1 and Relay 2.
    \item Scan attack: Launches a network scan using nmap. The attack is launched from the PC1 attack device. The attack targets Relay 1 and Relay 2. Two attacks are launched. The first attack consists of a nmap OS fingerprint scan. The second attack launches a TCP SYN port scan.
    \item Failed Authentication: An unauthorized user attempts to access remote devices, failing to get access after three attempts. The scenario is launch from PC2 and targets Relay 1 and Relay 2 through ssh and telnet. 
    After three unsuccessful login attempts, the relays pulse a series of alarms that are communicated through DNP3. 
    \item Setting change: An unauthorized user successfully accesses Relays 1 and 2 from PC2 and makes changes in the settings of each device.
    The setting changes include changing the current transformer ratio, phase instantaneous overcurrent level, and relay trip equations.
    When the user logs into the device, the relay pulses an alarm which is communicated through DNP3 to the RTAC device.
    After the setting change is executed, the relay pulses another alarm which is also communicated through DNP3.
    
\end{itemize}

\subsection{Global features: anomaly detection}

\begin{figure}[t]
    \centering
    \includegraphics[scale=0.5,trim={1.3cm 1.3cm 1.3cm 1.3cm},clip]{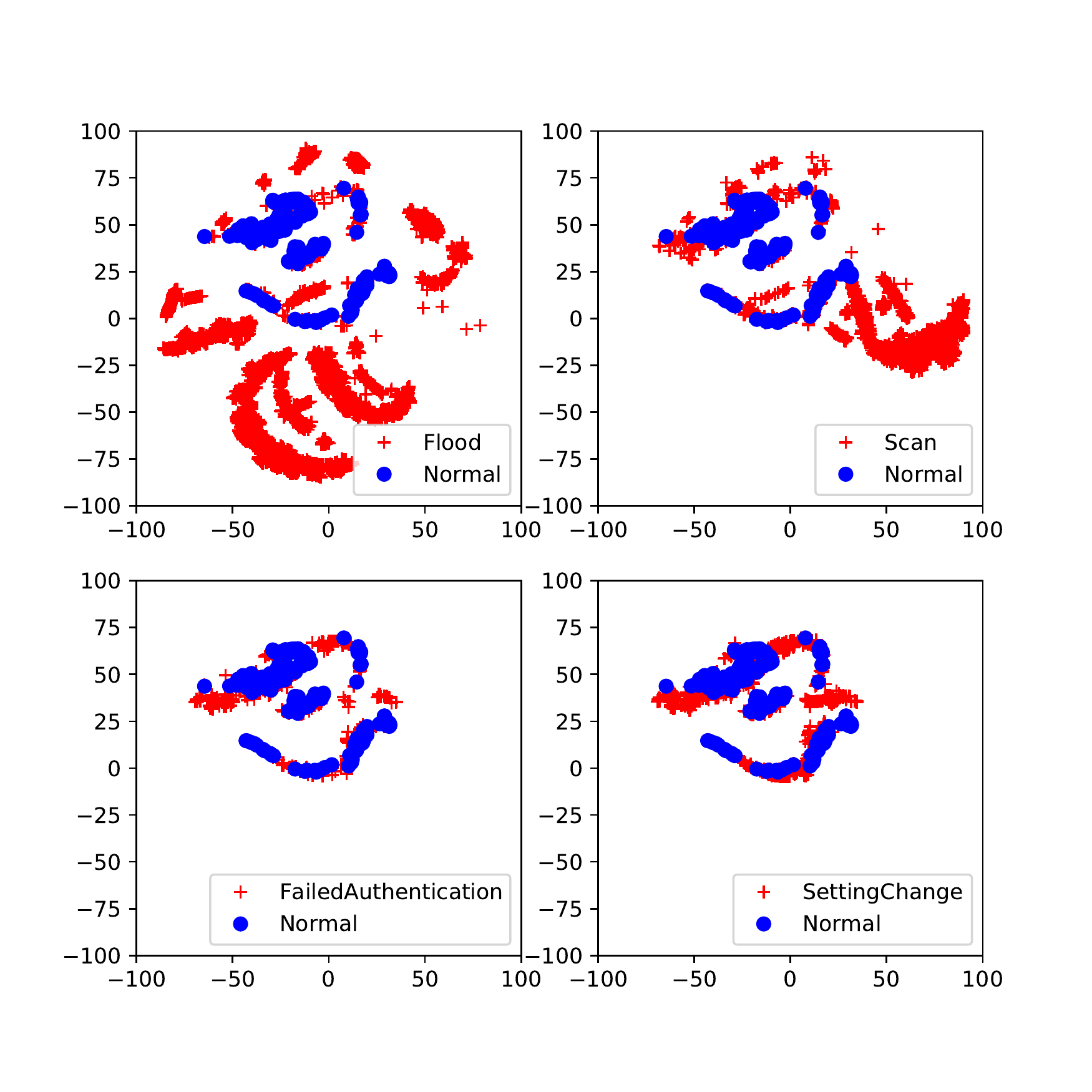}
    \caption{Global features visualized using the T-SNE algorithm}
    \label{fig:global_embeddings}
\end{figure}

\begin{table*}[t]
    \centering
    \scalebox{0.99}{
    \begin{tabular}{lrrr|rrrr}
    \toprule
    {} &  FPR train &  FPR test &       ADR &  Flood &  Failed Auth &  Scan &  Setting Change \\
    \midrule
    Baseline LOF   &   0.0931 &  0.1009 &  \textbf{0.7486} &   \textbf{0.8855} &                  \textbf{0.3791} &  \textbf{0.8439} &           \textbf{0.7063} \\
    Baseline OCSVM &   0.0998 &  0.1062 &  0.7023 &   0.8655 &                  0.3059 &  0.8105 &           0.6267 \\
    Baseline AE    &   \textbf{0.0092} &  \textbf{0.0363} &  0.6923 &   0.8648 &                  0.2934 &  0.7915 &           0.616148 \\
    \midrule
    NeT-LOF   &   0.0896 &  0.1010 &  \textbf{0.8232} &   0.9270 &                  \textbf{0.3677} &  0.8423 &           \textbf{0.6090} \\
    NeT-OCSVM &   0.1005 &  0.1046 &  0.5124 &   0.7954 &                  0.2495 &  0.1131 &           0.2732 \\
    NeT-AE    &   \textbf{0.0036} &  \textbf{0.0336} &  0.8249 &   \textbf{0.9609} &                  0.2848 &  \textbf{0.8492} &           0.5128 \\
    \midrule
    NeTB-LOF   &   0.0907 &  0.1019 &  \textbf{0.8471} &   0.9514 &                  \textbf{0.3465} &  \textbf{0.8634} &           \textbf{0.6634} \\
    NeTB-OCSVM &   0.1001 &  0.1001 &  0.2341 &   0.2805 &                  0.2495 &  0.1099 &           0.2804 \\
    NeTB-AE    &   \textbf{0.0097} &  \textbf{0.0327} &  0.8339 &   \textbf{0.9670} &                  0.1980 &  0.8539 &           0.599073 \\
    \bottomrule
    \end{tabular}
    }
    \caption{False Positive Rate (FPR) and Anomaly Detection Rate (ADR) for Network Transformer (NeT) model.}
    \label{tab:cybergraph_results}
\end{table*}

As described in section \ref{section:graph_features}, global features are used to characterize the behavior of the entire network.
Figure \ref{fig:global_embeddings} shows a visualization of the NeT global features.
The visualization is obtained with the T-SNE algorithm \cite{van2008visualizing}.
The figure shows each anomaly scenario (red) along with data from normal operations (blue).
The figure provides a visual reference of how different each scenario is with respect to the normal behavior of the system.
We observe that features from flood and scan scenarios are significantly different from normal behavior. 
Failed authentication and setting change have more subtle differences from normal data.
This behavior follows our expectations as flood and scan attacks have a large impact on the network as both introduce a large volume of packets.
This visual representation of the global features serves as an initial understanding of the data, providing a tool to understand the similarity between different scenarios.

Table \ref{tab:cybergraph_results} shows the anomaly detection results obtained with the NeT global features.
Three anomaly detection algorithms were considered in this analysis: Local Outlier Factor (LOF), One-Class SVM (OCSVM), and Autoencoder (AE).
These anomaly detection algorithms run on top of the NeT global features.
The baseline uses the same three anomaly detection algorithms, but it considers a series of hand-engineering statistical features found in the literature \cite{marino2021data,marino2019cps}. 
These features are computed across packet windows of 30 seconds.
NeT refers to the Network Transformer using the TCP features from Table \ref{tab:TCPFeatures}.
NeTB refers to the Network Transformer using the raw Bytes features from Table \ref{tab:ByteFeatures}.
NeT and NeTB report anomalies detected using packet windows of 30 seconds as well.
The AE model uses the reconstruction error to identify anomalies \cite{marino2021data}.
The table shows the performance measured using False Positive Ratio (FPR) and Anomaly Detection Ratio (ADR).
Results are measured using 5-fold cross-validation.
The FPR measures the rate of anomalies for normal scenarios.
The ADR measures the rate of anomalies during attack scenarios.
We report ADR measured across all scenarios and for each attack scenario separately.
Values of FPR closer to zero and high values of ADR indicate better anomaly detection performance. 
It is worth clarifying that the ADR may not necessarily be one, as the anomalous scenarios contain a mix of normal and abnormal data \cite{marino2021data}.

When looking at the overall results presented in Table \ref{tab:cybergraph_results}, AE provided the lowest FPR while achieving comparable ADR performance.
NeTB and NeT provided slightly better FPR with higher ADR when compared with the baseline.
We observed a higher ADR for Flood and Scan scenarios using NeTB and NeT models.
However, the baseline provided a slightly better performance on failed authentication and setting change scenarios.
NeT-AE and Baseline-AE provided comparable results for failed authentication, with NeTB-AE providing the lowest performance.
For setting change, NeTB-AE and Baseline AE provided comparable results, with NeT-AE having the lowest performance in this scenario.
The baseline and the NT models have specific features indicating activity on ssh and telnet, which helps to explain why they perform better in the failed authentication scenario.
On the other hand, setting change is characterized by larger payloads than failed authentication, which helps to explain the better performance of NeTB as it includes the raw 512 bytes of the payload.

Table \ref{tab:cybergraph_results} shows that the presented NeT and NeTB models provide comparable or better performance in anomaly detection than the baseline.
Furthermore, the baseline has no mechanisms to explore which devices and connections are involved in the anomalies.
The following experiments show the capability of NeT to provide an indication of devices affected and anomalous connections using the Node and Edge features.
Given the leading performance of AE over LOF and OCSVM, the following sections use AE for all experiments.

\subsection{Node features: Identify devices affected}
\begin{figure}[t]
    \centering
    \includegraphics[scale=0.45]{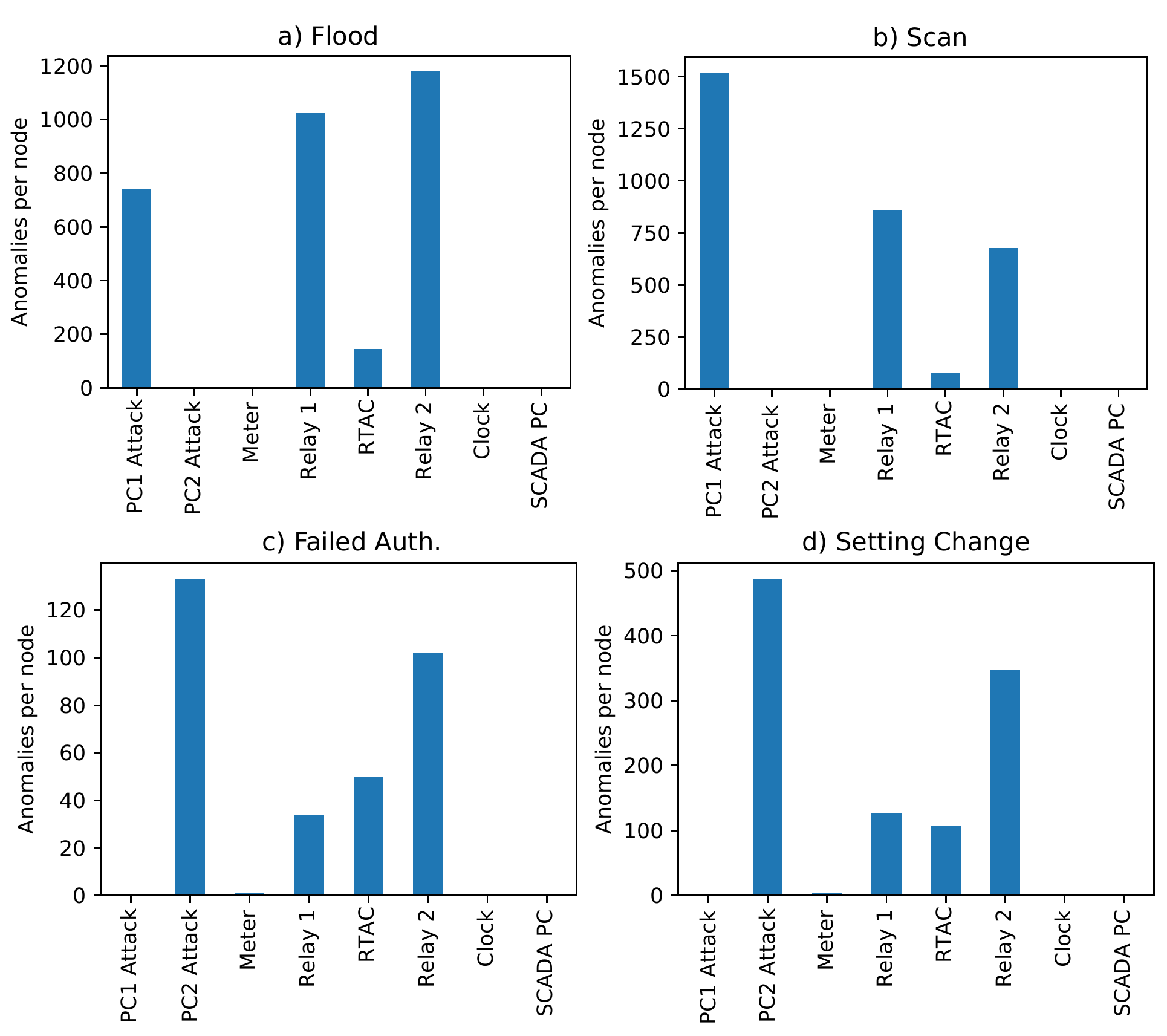}
    \caption{Anomalies in each device. Anomalies are obtained using an AE with NeT node features.}
    \label{fig:node_features}
\end{figure}

Node features are used to characterize the behavior of each device in the network.
These features can be used to identify the devices affected by an anomaly.
Figure \ref{fig:node_features} shows the results of devices affected obtained by analyzing the Node features.
The figure shows the number of anomalies per device for each one of the attack scenarios.

First, we observe that the presented approach is able to recognize which PC is launching the attack.
As described in section \ref{section:cyber_data_collection}, Flood and Scan attacks are launched by PC1 while Failed authentication and Setting Change are launched by PC2.
Figure \ref{fig:node_features} clearly shows a high anomaly rate for PC1 in Flood and Scan scenarios while PC2 shows no anomaly.
For Failed authentication and Setting Change, PC2 shows a high anomaly rate while PC1 shows no anomalies.
Figure \ref{fig:node_features} also shows a high anomaly ratio for Relay1 and Relay2. These relays are the devices targeted by the attacker, evidencing how the presented approach is able to detect the devices being targeted.

Figure \ref{fig:node_features} also shows a high number of anomalies for the RTAC device during Failed authentication and Setting Change.
Although the RTAC device was not targeted by the cyber attacks, the relays communicate with the RTAC whenever there is a failed authentication or a setting change.
This communication happens using the DNP3 protocol, and it is described in section \ref{section:cyber_data_collection}.
This is an interesting observation because it shows how the presented approach not only detects the attacker and target devices but also detects side effects from the attack.

\subsection{Edge features: Identify anomalous connections}
\begin{figure}[t]
    \centering
    \includegraphics[scale=0.49]{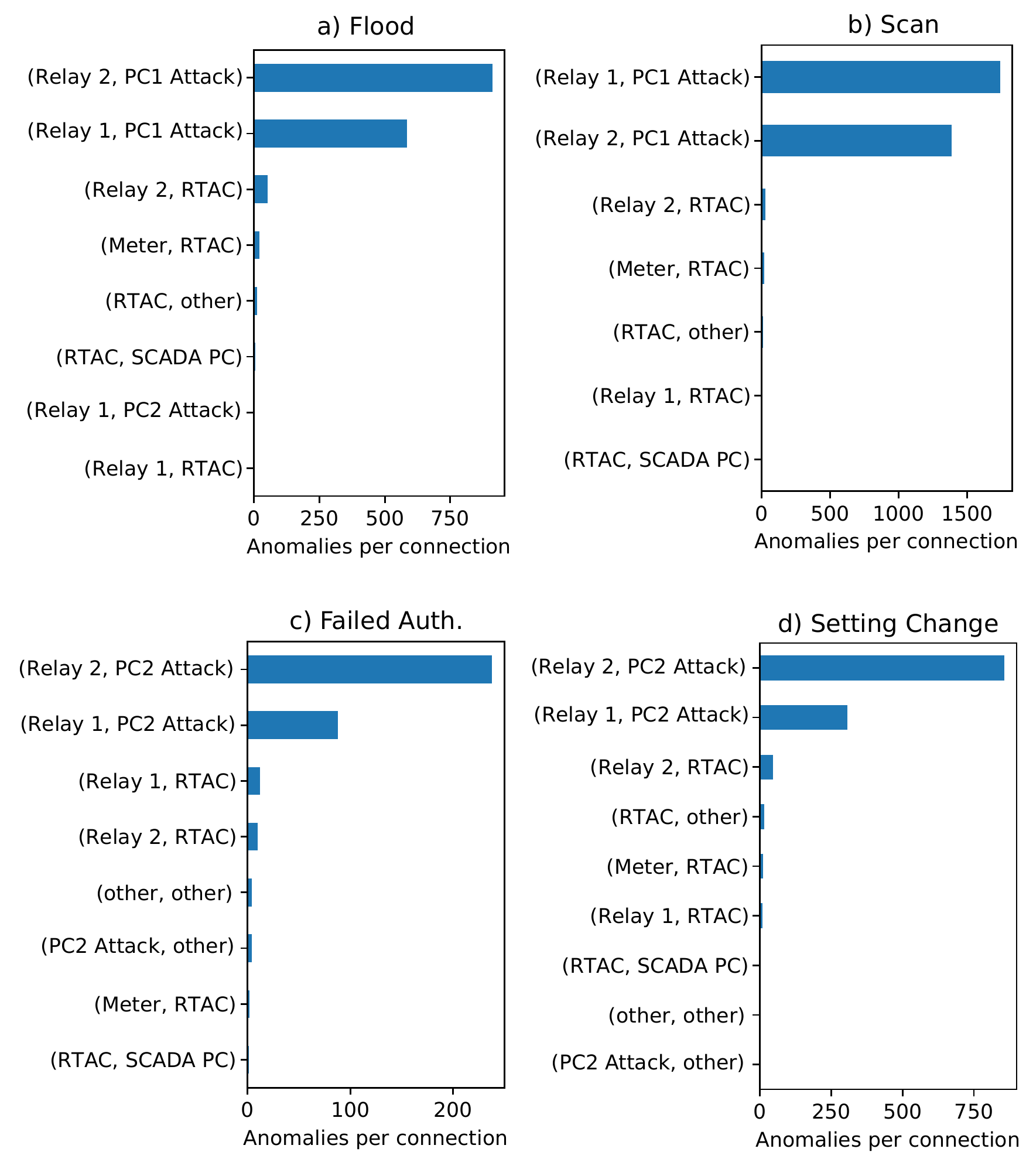}
    \caption{Anomalous connections inspected with the NeT model. Anomalies are detected using AE on NeT edge features.}
    \label{fig:edge_features}
\end{figure}

Edge features are used to characterize each connection between devices in the network.
By running anomaly detection on the Edge features, we are able to identify the connections responsible for an anomaly.
Figure \ref{fig:edge_features} shows the results of anomalous connections obtained by analyzing the Edge features.
The figure shows that the presented approach is able to identify the connections responsible for the anomalies.
For the Flood and Scan scenarios, we observe that the approach successfully reports the anomalous connections between the attacker (PC1) and the relays.
For Failed Authentication and Setting Change, the presented approach is able to identify that the anomalous connections involve PC2 and the relays.

Figures \ref{fig:global_embeddings}, \ref{fig:node_features}, and \ref{fig:edge_features}  demonstrate how the presented approach allows to extract information from several abstraction layers thanks to the graph structure embedded in the presented approach.
Thanks to the equivalence between the learned graph representation and the real device network communication, it is easy for an expert to exploit the learned model in order to extract useful and more detailed information.

  \section{Related work}
\label{section:cyber_related_work}

Anomaly detection refers to identifying patterns in data which does not confront the expected behavior of a system \cite{10.1145/1541880.1541882}. 
Anomaly detection can be addressed in many different forms such as out-of-distribution detection, outlier detection, and novelty detection \cite{NEURIPS2018_5e62d03a, 10.1145/3444690}.
A common approach for anomaly detection in computer networks is to use unsupervised ML models with features extracted from log files, including network traffic, kernel logs, SCADA logs, among others \cite{skopik2020synergy,zhao2021lstm}.
Hand engineered statistical and temporal features are also commonly used as an input to ML anomaly detection algorithms \cite{marino2017data,marino2021access,marino2021data}.

This paper used three unsupervised ML algorithms to perform anomaly detection from the Network Transformer hierarchical features.
These algorithms are One-Class Support Vector Machines (OCSVMs), Local Outlier Factor (LOFs), and Autoencoders (AEs). 
The main reason for using these three algorithms is that they do not require labeled data.
The labeling process is expensive, requires expertise on the data itself, and in the case of anomaly detection, it requires abnormal data, which is in many scenarios very hard to obtain \cite{langkvist2014review,wickramasinghe2021resnet,marino2021data}.
OCSVMs are extensions of Support Vector Machines (SVMs) \cite{scholkopf1999support,scholkopf2001estimating}. They have been successfully used in anomaly detection for network traffic \cite{8232594,marino2021access}.
LOF has also been extensively used in anomaly detection for cyber-physical systems \cite{sandhya2020comparative,ahmed2015investigation, breunig2000lof}.
Autoencoders (AEs) are a popular Neural Network architecture. They have been successfully used in several anomaly detection applications as well \cite{10.1145/3097983.3098052, wickramasinghe2018generalization}.

Self-supervised learning has increasingly gained attention for Out of Distribution (OoD) detection \cite{sohn2020learning,li2021cutpaste,georgescu2021anomaly}. 
OoD is a problem closely related to anomaly detection, where the objective is to detect samples that do not belong to the distribution of a given dataset.
One such example is presented in \cite{sohn2020learning}, where authors perform contrastive learning alongside OCSVM to detect OoD samples. 

Explainable AI is another area that has gained attention in recent years \cite{gunning2019darpa}.
LIME and SHAP \cite{ribeiro2016should,NIPS2017_7062} are popular algorithms that can be used to perform local explanations of Transformer models.
Incorporating domain knowledge in the inductive bias of ML algorithms is another approach to improve interpretability \cite{marino2021physics}
Graph nets provide a flexible approach to embed relational inductive bias in ML \cite{graphnets2018arxiv}.
Graph Neural Networks (GNNs) have been recently proposed to learn the structure of existing relationships between variables while performing anomaly detection in time-series sensor data \cite{deng2021graph}.
GNNs have also been proposed to identify anomalous edges in dynamic graphs \cite{cai2021structural}.

\section{Conclusion}
\label{section:cyber_conclusion}

This paper presented the Network Transformer, a ML model that uses graph-based representations to incorporate the structure of a monitored computer network.
The presented model uses self-supervised learning in order to train the model without the need for labeled data.
The Network Transformer provides an approach to extract a series of hierarchical graph features for down-stream ML analysis.
In this paper, we demonstrated the use of the extracted hierarchical features for anomaly detection.
The Network Transformer successfully identified anomalies in an ICS network while being able to report devices affected and connections compromised, 
demonstrating its ability for enhanced interpretability by facilitating the extraction of more detailed information about the anomalies.

The presented approach provided an end-to-end differentiable model for analysing network packet data, starting from potentially raw byte values up to a global representation of the network graph.
Although we used a complex DNN Transformer, the hierarchical design of the approach allows to backtrack and analyze the predictions of the model in different levels of granularity.
In the experimental evaluation we demonstrated this analysis starting from a global representation followed by an analysis of individual nodes and connections.
The presented analysis demonstrates how the graph structure can be exploited to not only identify anomalies but extract useful information of what devices the anomaly is affecting and which connections are responsible for the anomaly.
Because the Transformer Encoder processes potential raw binary data, the approach can be used as a unified end-to-end methodology for network traffic monitoring that goes from a global view up to individual bytes.
Future work will explore the use of existing explainable algorithms such as LIME or SHAP to provide explanations at the packet and possibly byte level.

\bibliographystyle{IEEEtran}

\begin{thebibliography}{10}
\providecommand{\url}[1]{#1}
\csname url@samestyle\endcsname
\providecommand{\newblock}{\relax}
\providecommand{\bibinfo}[2]{#2}
\providecommand{\BIBentrySTDinterwordspacing}{\spaceskip=0pt\relax}
\providecommand{\BIBentryALTinterwordstretchfactor}{4}
\providecommand{\BIBentryALTinterwordspacing}{\spaceskip=\fontdimen2\font plus
\BIBentryALTinterwordstretchfactor\fontdimen3\font minus
  \fontdimen4\font\relax}
\providecommand{\BIBforeignlanguage}[2]{{%
\expandafter\ifx\csname l@#1\endcsname\relax
\typeout{** WARNING: IEEEtran.bst: No hyphenation pattern has been}%
\typeout{** loaded for the language `#1'. Using the pattern for}%
\typeout{** the default language instead.}%
\else
\language=\csname l@#1\endcsname
\fi
#2}}
\providecommand{\BIBdecl}{\relax}
\BIBdecl

\bibitem{marino2021access}
D.~L. Marino, C.~S. Wickramasinghe, V.~K. Singh, J.~Gentle, C.~Rieger, and
  M.~Manic, ``The virtualized cyber-physical testbed for machine learning
  anomaly detection: A wind powered grid case study,'' \emph{IEEE Access},
  vol.~9, pp. 159\,475--159\,494, 2021.

\bibitem{en14051409}
\BIBentryALTinterwordspacing
B.~Vaagensmith, V.~K. Singh, R.~Ivans, D.~L. Marino, C.~S. Wickramasinghe,
  J.~Lehmer, T.~Phillips, C.~Rieger, and M.~Manic, ``Review of design elements
  within power infrastructure cyber–physical test beds as threat analysis
  environments,'' \emph{Energies}, vol.~14, no.~5, 2021. [Online]. Available:
  \url{https://www.mdpi.com/1996-1073/14/5/1409}
\BIBentrySTDinterwordspacing

\bibitem{marino2019cyber}
D.~Marino, C.~Wickramasinghe, K.~Amarasinghe, H.~Challa, P.~Richardson,
  A.~Jillepalli, B.~Johnson, C.~Rieger, and M.~Manic, ``Cyber and physical
  anomaly detection insmart-grids,'' in \emph{IEEE Resilience Week (RW)
  2019}.\hskip 1em plus 0.5em minus 0.4em\relax ACM, 2019.

\bibitem{marino2021data}
D.~L. Marino, C.~S. Wickramasinghe, B.~Tsouvalas, C.~Rieger, and M.~Manic,
  ``Data-driven correlation of cyber and physical anomalies for holistic system
  health monitoring,'' in \emph{IEEE ACCESS}.\hskip 1em plus 0.5em minus
  0.4em\relax IEEE, 2021.

\bibitem{marino2019data}
D.~L. Marino, C.~S. Wickramasinghe, C.~Rieger, and M.~Manic, ``Data-driven
  stochastic anomaly detection on smart-grid communications using mixture
  poisson distributions,'' in \emph{IECON 2019-45th Annual Conference of the
  IEEE Industrial Electronics Society}, vol.~1.\hskip 1em plus 0.5em minus
  0.4em\relax IEEE, 2019, pp. 5855--5861, \copyright 2019 IEEE.

\bibitem{zhao2021lstm}
Z.~Zhao, C.~Xu, and B.~Li, ``A lstm-based anomaly detection model for log
  analysis,'' \emph{Journal of Signal Processing Systems}, vol.~93, no.~7, pp.
  745--751, 2021.

\bibitem{skopik2020synergy}
F.~Skopik, M.~Landauer, M.~Wurzenberger, G.~Vormayr, J.~Milosevic, J.~Fabini,
  W.~Pr{\"u}ggler, O.~Kruschitz, B.~Widmann, K.~Truckenthanner \emph{et~al.},
  ``synergy: Cross-correlation of operational and contextual data to timely
  detect and mitigate attacks to cyber-physical systems,'' \emph{Journal of
  Information Security and Applications}, vol.~54, p. 102544, 2020.

\bibitem{gunning2017explainable}
D.~Gunning, ``Explainable artificial intelligence (xai),'' \emph{Defense
  Advanced Research Projects Agency (DARPA), nd Web}, vol.~2, 2017.

\bibitem{NIPS2017_Transformer}
\BIBentryALTinterwordspacing
A.~Vaswani, N.~Shazeer, N.~Parmar, J.~Uszkoreit, L.~Jones, A.~N. Gomez, L.~u.
  Kaiser, and I.~Polosukhin, ``Attention is all you need,'' in \emph{Advances
  in Neural Information Processing Systems}, I.~Guyon, U.~V. Luxburg,
  S.~Bengio, H.~Wallach, R.~Fergus, S.~Vishwanathan, and R.~Garnett, Eds.,
  vol.~30.\hskip 1em plus 0.5em minus 0.4em\relax Curran Associates, Inc.,
  2017. [Online]. Available:
  \url{https://proceedings.neurips.cc/paper/2017/file/3f5ee243547dee91fbd053c1c4a845aa-Paper.pdf}
\BIBentrySTDinterwordspacing

\bibitem{graphnets2018arxiv}
\BIBentryALTinterwordspacing
P.~Battaglia, J.~B.~C. Hamrick, V.~Bapst, A.~Sanchez, V.~Zambaldi,
  M.~Malinowski, A.~Tacchetti, D.~Raposo, A.~Santoro, R.~Faulkner, C.~Gulcehre,
  F.~Song, A.~Ballard, J.~Gilmer, G.~E. Dahl, A.~Vaswani, K.~Allen, C.~Nash,
  V.~J. Langston, C.~Dyer, N.~Heess, D.~Wierstra, P.~Kohli, M.~Botvinick,
  O.~Vinyals, Y.~Li, and R.~Pascanu, ``Relational inductive biases, deep
  learning, and graph networks,'' \emph{arXiv}, 2018. [Online]. Available:
  \url{https://arxiv.org/pdf/1806.01261.pdf}
\BIBentrySTDinterwordspacing

\bibitem{wang2018non}
X.~Wang, R.~Girshick, A.~Gupta, and K.~He, ``Non-local neural networks,'' in
  \emph{Proceedings of the IEEE conference on computer vision and pattern
  recognition}, 2018, pp. 7794--7803.

\bibitem{liu2021self}
X.~Liu, F.~Zhang, Z.~Hou, L.~Mian, Z.~Wang, J.~Zhang, and J.~Tang,
  ``Self-supervised learning: Generative or contrastive,'' \emph{IEEE
  Transactions on Knowledge and Data Engineering}, 2021.

\bibitem{he2016deep}
K.~He, X.~Zhang, S.~Ren, and J.~Sun, ``Deep residual learning for image
  recognition,'' in \emph{Proceedings of the IEEE conference on computer vision
  and pattern recognition}, 2016, pp. 770--778.

\bibitem{ba2016layer}
J.~L. Ba, J.~R. Kiros, and G.~E. Hinton, ``Layer normalization,'' \emph{arXiv
  preprint arXiv:1607.06450}, 2016.

\bibitem{kingma2014adam}
D.~P. Kingma and J.~Ba, ``Adam: {A} method for stochastic optimization,'' in
  \emph{3rd International Conference on Learning Representations ({ICLR})}, May
  2015.

\bibitem{van2008visualizing}
L.~Van~der Maaten and G.~Hinton, ``Visualizing data using t-sne.''
  \emph{Journal of machine learning research}, vol.~9, no.~11, 2008.

\bibitem{marino2019cps}
D.~L. Marino, C.~S. Wickramasinghe, K.~Amarasinghe, H.~Challa, P.~Richardson,
  A.~A. Jillepalli, B.~K. Johnson, C.~Rieger, and M.~Manic, ``Cyber and
  physical anomaly detection in smart-grids,'' in \emph{2019 Resilience Week
  (RWS)}, vol.~1, 2019, pp. 187--193.

\bibitem{10.1145/1541880.1541882}
\BIBentryALTinterwordspacing
V.~Chandola, A.~Banerjee, and V.~Kumar, ``Anomaly detection: A survey,''
  \emph{ACM Comput. Surv.}, vol.~41, no.~3, jul 2009. [Online]. Available:
  \url{https://doi.org/10.1145/1541880.1541882}
\BIBentrySTDinterwordspacing

\bibitem{NEURIPS2018_5e62d03a}
\BIBentryALTinterwordspacing
I.~Golan and R.~El-Yaniv, ``Deep anomaly detection using geometric
  transformations,'' in \emph{Advances in Neural Information Processing
  Systems}, S.~Bengio, H.~Wallach, H.~Larochelle, K.~Grauman, N.~Cesa-Bianchi,
  and R.~Garnett, Eds., vol.~31.\hskip 1em plus 0.5em minus 0.4em\relax Curran
  Associates, Inc., 2018. [Online]. Available:
  \url{https://proceedings.neurips.cc/paper/2018/file/5e62d03aec0d17facfc5355dd90d441c-Paper.pdf}
\BIBentrySTDinterwordspacing

\bibitem{10.1145/3444690}
\BIBentryALTinterwordspacing
A.~Bl\'{a}zquez-Garc\'{\i}a, A.~Conde, U.~Mori, and J.~A. Lozano, ``A review on
  outlier/anomaly detection in time series data,'' \emph{ACM Comput. Surv.},
  vol.~54, no.~3, apr 2021. [Online]. Available:
  \url{https://doi.org/10.1145/3444690}
\BIBentrySTDinterwordspacing

\bibitem{marino2017data}
D.~{Marino}, K.~{Amarasinghe}, M.~{Anderson}, N.~{Yancey}, Q.~{Nguyen},
  K.~{Kenney}, and M.~{Manic}, ``Data driven decision support for reliable
  biomass feedstock preprocessing,'' in \emph{2017 Resilience Week (RWS)}, Sep.
  2017, pp. 97--102.

\bibitem{langkvist2014review}
M.~L{\"a}ngkvist, L.~Karlsson, and A.~Loutfi, ``A review of unsupervised
  feature learning and deep learning for time-series modeling,'' \emph{Pattern
  Recognition Letters}, vol.~42, pp. 11--24, 2014.

\bibitem{wickramasinghe2021resnet}
C.~S. Wickramasinghe, D.~L. Marino, and M.~Manic, ``Resnet autoencoders for
  unsupervised feature learning from high-dimensional data: Deep models
  resistant to performance degradation,'' \emph{IEEE Access}, vol.~9, pp.
  40\,511--40\,520, 2021.

\bibitem{scholkopf1999support}
B.~Sch{\"o}lkopf, R.~C. Williamson, A.~J. Smola, J.~Shawe-Taylor, J.~C. Platt
  \emph{et~al.}, ``Support vector method for novelty detection.'' in
  \emph{NIPS}, vol.~12.\hskip 1em plus 0.5em minus 0.4em\relax Citeseer, 1999,
  pp. 582--588.

\bibitem{scholkopf2001estimating}
B.~Sch{\"o}lkopf, J.~C. Platt, J.~Shawe-Taylor, A.~J. Smola, and R.~C.
  Williamson, ``Estimating the support of a high-dimensional distribution,''
  \emph{Neural computation}, vol.~13, no.~7, pp. 1443--1471, 2001.

\bibitem{8232594}
S.~{Teng}, N.~{Wu}, H.~{Zhu}, L.~{Teng}, and W.~{Zhang}, ``Svm-dt-based
  adaptive and collaborative intrusion detection,'' \emph{IEEE/CAA Journal of
  Automatica Sinica}, vol.~5, no.~1, pp. 108--118, 2018.

\bibitem{sandhya2020comparative}
R.~Sandhya, J.~Prakash, and B.~V. Kumar, ``Comparative analysis of clustering
  techniques in anomaly detection wind turbine data.'' \emph{Journal of Xi’an
  University of Architecture \& Technology}, vol.~12, no.~3, pp. 5684--5694,
  2020.

\bibitem{ahmed2015investigation}
M.~Ahmed, A.~Anwar, A.~N. Mahmood, Z.~Shah, and M.~J. Maher, ``An investigation
  of performance analysis of anomaly detection techniques for big data in scada
  systems.'' \emph{EAI Endorsed Trans. Indust. Netw. \& Intellig. Syst.},
  vol.~2, no.~3, p.~e5, 2015.

\bibitem{breunig2000lof}
M.~M. Breunig, H.-P. Kriegel, R.~T. Ng, and J.~Sander, ``Lof: identifying
  density-based local outliers,'' in \emph{Proceedings of the 2000 ACM SIGMOD
  international conference on Management of data}, 2000, pp. 93--104.

\bibitem{10.1145/3097983.3098052}
C.~Zhou and R.~C. Paffenroth, ``Anomaly detection with robust deep
  autoencoders,'' in \emph{Proceedings of the 23rd ACM SIGKDD International
  Conference on Knowledge Discovery and Data Mining}, ser. {KDD '17}.\hskip 1em
  plus 0.5em minus 0.4em\relax New York, NY, USA: Association for Computing
  Machinery, 2017, p. 665–674.

\bibitem{wickramasinghe2018generalization}
C.~S. Wickramasinghe, D.~L. Marino, K.~Amarasinghe, and M.~Manic,
  ``Generalization of deep learning for cyber-physical system security: A
  survey,'' in \emph{IECON 2018-44th Annual Conference of the IEEE Industrial
  Electronics Society}.\hskip 1em plus 0.5em minus 0.4em\relax IEEE, 2018, pp.
  745--751.

\bibitem{sohn2020learning}
K.~Sohn, C.-L. Li, J.~Yoon, M.~Jin, and T.~Pfister, ``Learning and evaluating
  representations for deep one-class classification,'' \emph{arXiv preprint
  arXiv:2011.02578}, 2020.

\bibitem{li2021cutpaste}
C.-L. Li, K.~Sohn, J.~Yoon, and T.~Pfister, ``Cutpaste: Self-supervised
  learning for anomaly detection and localization,'' in \emph{Proceedings of
  the IEEE/CVF Conference on Computer Vision and Pattern Recognition}, 2021,
  pp. 9664--9674.

\bibitem{georgescu2021anomaly}
M.-I. Georgescu, A.~Barbalau, R.~T. Ionescu, F.~S. Khan, M.~Popescu, and
  M.~Shah, ``Anomaly detection in video via self-supervised and multi-task
  learning,'' in \emph{Proceedings of the IEEE/CVF Conference on Computer
  Vision and Pattern Recognition}, 2021, pp. 12\,742--12\,752.

\bibitem{gunning2019darpa}
D.~Gunning and D.~Aha, ``Darpa’s explainable artificial intelligence (xai)
  program,'' \emph{AI Magazine}, vol.~40, no.~2, pp. 44--58, 2019.

\bibitem{ribeiro2016should}
M.~T. Ribeiro, S.~Singh, and C.~Guestrin, ``" why should i trust you?"
  explaining the predictions of any classifier,'' in \emph{Proceedings of the
  22nd ACM SIGKDD international conference on knowledge discovery and data
  mining}, 2016, pp. 1135--1144.

\bibitem{NIPS2017_7062}
\BIBentryALTinterwordspacing
S.~M. Lundberg and S.-I. Lee, ``A unified approach to interpreting model
  predictions,'' in \emph{Advances in Neural Information Processing Systems
  30}, I.~Guyon, U.~V. Luxburg, S.~Bengio, H.~Wallach, R.~Fergus,
  S.~Vishwanathan, and R.~Garnett, Eds.\hskip 1em plus 0.5em minus 0.4em\relax
  Curran Associates, Inc., 2017, pp. 4765--4774. [Online]. Available:
  \url{http://papers.nips.cc/paper/7062-a-unified-approach-to-interpreting-model-predictions.pdf}
\BIBentrySTDinterwordspacing

\bibitem{marino2021physics}
{{\textcopyright}[2021] IEEE. Reprinted, with permission from D. L. Marino} and
  M.~{Manic}, ``Physics enhanced data-driven models with variational gaussian
  processes,'' \emph{IEEE Open Journal of the Industrial Electronics Society},
  vol.~2, pp. 252--265, 2021.

\bibitem{deng2021graph}
A.~Deng and B.~Hooi, ``Graph neural network-based anomaly detection in
  multivariate time series,'' in \emph{Proceedings of the AAAI Conference on
  Artificial Intelligence}, vol.~35, no.~5, 2021, pp. 4027--4035.

\bibitem{cai2021structural}
L.~Cai, Z.~Chen, C.~Luo, J.~Gui, J.~Ni, D.~Li, and H.~Chen, ``Structural
  temporal graph neural networks for anomaly detection in dynamic graphs,'' in
  \emph{Proceedings of the 30th ACM International Conference on Information \&
  Knowledge Management}, 2021, pp. 3747--3756.

\end{thebibliography}

\end{document}